\documentclass[letterpaper, 10 pt, conference]{ieeeconf}  % Comment this line out if you need a4paper

\IEEEoverridecommandlockouts                              % This command is only needed if 
\usepackage{epsfig} % for postscript graphics files
\usepackage{mathptmx} % assumes new font selection scheme installed
\usepackage{times} % assumes new font selection scheme installed
\usepackage{amsmath} % assumes amsmath package installed
\usepackage{amssymb}  % assumes amsmath package installed
\usepackage{graphicx}

\title{\LARGE \bf
A Multi-State Social Force Based Framework for Vehicle-Pedestrian Interaction in Uncontrolled Pedestrian Crossing Scenarios
}

\author{Dongfang Yang$^{1}$, Keith Redmill$^{1}$, \IEEEmembership{IEEE~Senior~Member}, and \"{U}mit~\"{O}zg\"{u}ner$^{1}$, \IEEEmembership{IEEE~Life~Fellow~}% <-this % stops a space
\thanks{Material reported here was supported by the United States Department of Transportation under Award Number 69A3551747111 for the Mobility21 University Transportation Center.}% <-this % stops a space
\thanks{$^{1}$ Dongfang Yang, Keith Redmill and \"{U}mit~\"{O}zg\"{u}ner are with Electrical and Computer Engineering, The Ohio State University, Columbus, OH, USA. {\tt\small\{yang.3455,redmill.1,ozguner.1\}@osu.edu}}%
\thanks{$^{2}$ https://github.com/dongfang-steven-yang/vpi-crossing}%
}

\begin{document}

\maketitle
\thispagestyle{empty}
\pagestyle{empty}

%%%%%%%%%%%%%%%%%%%%%%%%%%%%%%%%%%%%%%%%%%%%%%%%%%%%%%%%%%%%%%%%%%%%%%%%%%%%%%%%
\begin{abstract}
Vehicle-pedestrian interaction (VPI) is one of the most challenging tasks for automated driving systems. The design of driving strategies for such systems usually starts with verifying VPI in simulation. This work proposed an improved framework for the study of VPI in uncontrolled pedestrian crossing scenarios. The framework admits the mutual effect between the pedestrian and the vehicle.
A multi-state social force based pedestrian motion model was designed to describe the microscopic motion of the pedestrian crossing behavior. The pedestrian model considers major interaction factors such as the accepted gap of the pedestrian's decision on when to start crossing, the desired speed of the pedestrian, and the effect of the vehicle on the pedestrian while the pedestrian is crossing the road. Vehicle driving strategies focus on the longitudinal motion control, for which the feedback obstacle avoidance control and the model predictive control were tested and compared in the framework. The simulation results$^{2}$ verified that the proposed framework can generate a variety of VPI scenarios, consisting of either the pedestrian yielding to the vehicle or the vehicle yielding to the pedestrian. The framework can be easily extended to apply different approaches to the VPI problems.

\end{abstract}

%%%%%%%%%%%%%%%%%%%%%%%%%%%%%%%%%%%%%%%%%%%%%%%%%%%%%%%%%%%%%%%%%%%%%%%%%%%%%%%%
\section{INTRODUCTION}

% Consider adding this reference: gorrini2018observation

\begin{figure}
    \centering
    \includegraphics[width=0.8\linewidth]{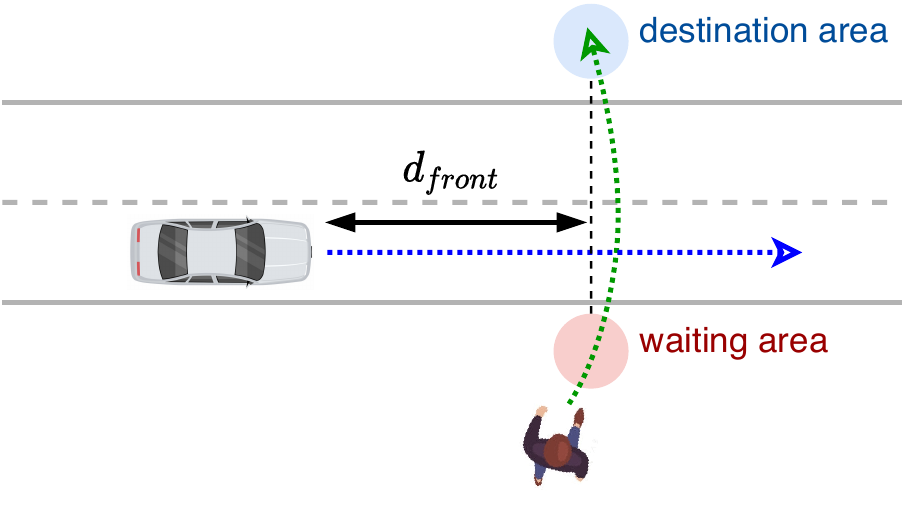}
    \caption{Scenario illustration. The pedestrian first goes to the waiting area (red circle), judges the situation to decide crossing or yielding, and walks to the destination area (blue circle).
    % The pedestrian motion follows transitions of approaching the waiting area (red circle), deciding to cross or yield to the vehicle at the waiting area, and crossing the road to reach the destination area (blue circle). In the process of crossing, the pedestrian behavior (speed, acceleration, and steering) is affected by the interacting vehicle, hence producing realistic and possibly non-linear motion. 
    The vehicle regulates it longitudinal speed to balance between the safety and the efficiency of finishing the interaction.}
    \label{fig:scenario_crossing}
\end{figure}

Pedestrian safety has been an important issue in transportation for a long time. According to NHTSA~\cite{nhtsa2019pedestrian}, the percentage of pedestrian fatalities in total traffic fatalities has increased from 12\% in 2008 to 16\% in 2017. This implies that pedestrian safety is still a big concern. Developing advanced driver-assistance systems (ADAS) is a promising route to improve pedestrian safety. Although research, applications, and products are continuously updating, the approaches for the vehicle to handle the vehicle-pedestrian interaction (VPI) still need to be improved, especially in uncontrolled scenarios (no crosswalk and no traffic signals). For example, a recent report by NTSB~\cite{ntsb_report} about a pedestrian jaywalking fatality involved with an autonomously driving vehicle demonstrated the inadequacy of the pedestrian handling functionality in the automated driving system. 
% The verification of such functionality requires trial and error hence an effective framework of VPI simulation is necessary. 
Therefore, this work focuses on the VPI scenario in which an autonomous vehicle interacts with a crossing pedestrian at uncontrolled road segments, as illustrated in Fig.\ref{fig:scenario_crossing}. This scenario is very common and is closely linked to pedestrian safety. 

It is generally challenging to model and evaluate such VPI because various VPI patterns can not be usually observed in real-world situations. Having an effective VPI framework for simulation would benefit the testing of newly designed algorithms before moving to the next step. To this end, this work proposed an improved VPI framework that can produce more realistic pedestrian behaviors by extending the social force pedestrian motion model~\cite{yang2020social}. The framework is suitable for evaluating different automated driving strategies in different situations.

% general study of vehicle-pedestrian interaction
Several works have studied the VPI in crossing scenarios. A recent comprehensive review~\cite{rasouli2019autonomous} identified factors such as pedestrian demographics, traffic dynamics, and environmental conditions by surveying both the classical driver-pedestrian interaction and the VPI that involves automated vehicles. Gap acceptance is a major factor that affects the pedestrian's behavior in crossing scenarios. Existing works like~\cite{woodman2019gap,yannis2013pedestrian} studied the gap acceptance in different conditions, i.e., at mid-blocks and in front of a platooning of low-speed autonomous pods. 
A stochastic interaction model using the multivariate Gaussian mixture model~\cite{chen2017evaluation} was also proposed to simulate the mutual interaction by simultaneously considering the behavior of both the vehicle and the pedestrian. The above works addressed the VPI in a statistic way, however, if we focus on the precise motion of the interacting agents, it is expected to have a more detailed VPI framework that models both agents' dynamics.

% vehicle control strategies
Regulating longitudinal speed is the most direct approach for the vehicle to handle the VPI in crossing scenarios. 
Partially observable Markov decision process (POMDP)~\cite{bandyopadhyay2013intention, thornton2018value, schraner2019pedestrian} is one of the popular methods for discretized longitudinal control incorporating the uncertainty of the pedestrian behavior. Model-based control methods like hybrid feedback control~\cite{ kapania2019hybrid} and model predictive control~\cite{liu2015predictive} are also suitable for this type of problem. Although model-based control does not inherently consider the pedestrian's behavior, a prediction module~\cite{goldhammer2019intentions} can be applied to provide the predicted pedestrian motion hence utilized by the model-based controllers.

% behavior of pedestrians
Pedestrian behavior is usually described by a motion model that can execute decisions like when and how to cross the road. A common assumption is that the pedestrian simply makes the decision on when to start crossing but while the pedestrian is crossing, a constant crossing velocity is maintained~\cite{chen2017evaluation, kapania2019hybrid}. 
The assumption is good for analyzing decision-making, but not for generating precise motion.
This work introduced a new model that combines the microscopic social force pedestrian model with a state machine to describe the explicit motion of the pedestrian in crossing scenarios. 
Social force model~\cite{helbing1995social} was originally designed for simulating the crowd dynamics of multiple interacting pedestrians. Recently, the effect of the vehicle on pedestrians has been added into the social force model~\cite{zeng2014application, yang2018social, yang2020social}, which makes it possible to be used in the VPI framework. In terms of decision making, we designed a state machine to handle different phases of a complete crossing behavior, which includes approaching the road, judging the situation, and crossing the road. 

The contribution of this work is summarized as follows: (a) A general VPI framework for uncontrolled crossing scenarios was proposed and can simulate various VPI patterns. Both the vehicle and the pedestrian have a hierarchical pipeline of perception, interaction, and motion that interact with each other. (b) A newly designed multi-state social force model was proposed to describe more realistic pedestrian motion under the effect of the interacting vehicle, i.e., the pedestrian can change velocity and direction in the process of crossing. The proposed pedestrian model also introduces the uncertainty in both the gap acceptance and the desired crossing speed. (c) Different vehicle control strategies (pure velocity keeping, obstacle avoidance, and model predictive) were implemented to verify the effectiveness and the capability of the VPI framework. The simulation can successfully generate various VPI patterns in the uncontrolled crossing scenarios.

% how to obtain pedestrian motion: 
% This work proposed a simulation framework to facilitate the study of the pedestrian crossing interaction. Therefore, the actual motion of the pedestrian is simulated by social force model that includes the interaction with vehicle. The simulated pedestrian motion is assumed to be accurate enough. Only the observation, i.e., current and past states of the pedestrian, is available to the vehicle, the vehicle would never know the mechanism behind the observation. The vehicle should take advantage of the observation and perform necessary inference to predict the future intention/motion of the pedestrian. 

% --------------------------------------------------------------------------------
\section{FRAMEWORK}
\begin{figure}
    \centering
    \includegraphics[width=\linewidth]{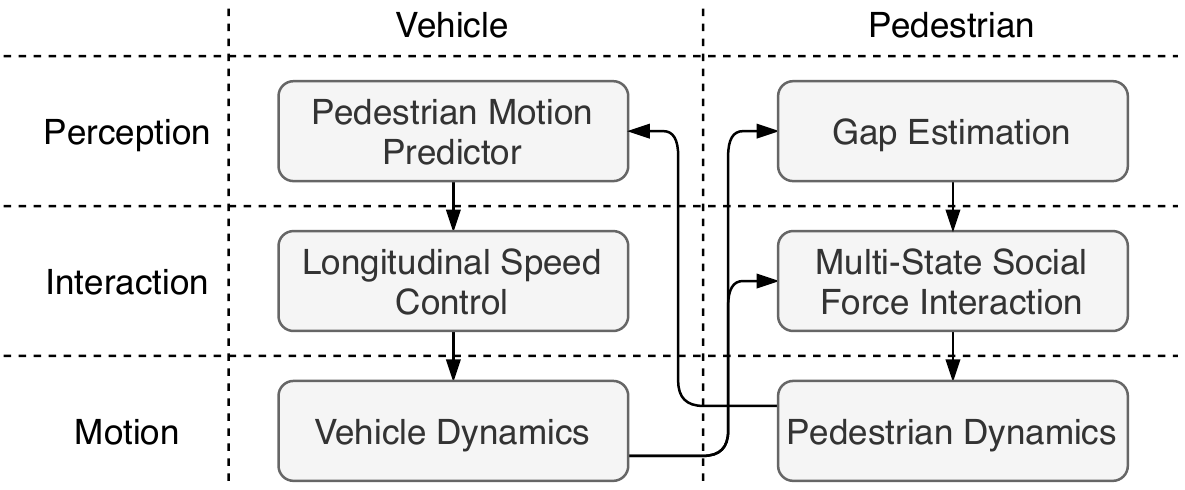}
    \caption{Framework for vehicle-pedestrian interaction. Both the vehicle and the pedestrian follows a 3-level hierarchy.}
    \label{fig:framework}
\end{figure}
The process of VPI can be interpreted as a process of two agents mutually recognizing and affecting each other. A general framework consisting of the same hierarchy for either the pedestrian or the vehicle can be conceptually divided as layers of perception, interaction, and motion, as illustrated in Fig.~\ref{fig:framework}. All of these layers should ideally interact with any of the others.
% The perception layer is responsible for the detection, recognition, and prediction of the interacting agent. The interaction layer produces the interactive decisions and generates the action that changes the agent's motion. And finally, the motion layer executes the action with physics-based dynamics. Ideally, all layers of both agents can interact with any other layer. 

In this work, we are dealing with the VPI in a specific and representative scenario as shown in Fig.~\ref{fig:scenario_crossing}. The vehicle moves along the lane adjacent to the side where the pedestrian appears, predicts whether the pedestrian is going to cross the road, and adjusts its speed accordingly to avoid the collision while keeping its desired speed as much as possible. The pedestrian emerges from the sidewalk, goes to the waiting area, and judges the situation to see if it is safe to cross the road. If safe the pedestrian will proceed to cross, otherwise, the pedestrian will wait until the situation becomes safe.

To address this specific scenario, the interaction among the aforementioned layers was streamlined as described by the arrows in Fig.~\ref{fig:framework}. Both the vehicle and the pedestrian are assumed to know the exact past and current states of each other. Therefore, the vehicle's perception layer employs a predictor to predict the pedestrian's future motion. And the pedestrian's perception layer applies an estimator for the time gap of the vehicle going from its real-time position to the pedestrian's crossing position. The interaction is not controlled by either traffic signals or crosswalk markings, which represents the common situations in residential areas or when the pedestrian jaywalks. The vehicle's interaction layer adopts a longitudinal speed control that can somehow take advantage of the trajectory predicted by the pedestrian motion predictor. The pedestrian's interaction layer uses the newly-designed multi-state social force model that allows the pedestrian to change the speed and the direction to avoid uncomfortable movement due to the approaching vehicle.
%which allows the pedestrian to change the speed while crossing the road and to slightly adjust the walking direction to avoid uncomfortable movement in the cases where the vehicle is pretty close to the pedestrian. 
Both bottom layers apply the dynamics to obey the physics of the motion. All the above layers are explained in detail in the following sections.

% \begin{figure}
%     \centering
%     \includegraphics[width=0.8\linewidth]{figures/framework_vpi.pdf}
%     \caption{Proposed framework for vehicle-pedestrian interaction in crossing scenarios.}
%     \label{fig:framework_vpi}
% \end{figure}

% --------------------------------------------------------------------------------
\section{INTERACTIVE PEDESTRIAN MOTION}
% This work designed a social force based pedestrian motion model that combines the state machine to describe the pedestrian motion while crossing the road.
% The pedestrian motion is mainly  state machine governs high-level decisions while the social force describes the microscopic motion. The pedestrian can adjust walking speed if necessary. For example, the pedestrian can accelerate in the middle of the crossing if the vehicle does not slow down. 

\subsection{Gap Estimation}
As discussed in~\cite{schmidt2009pedestrians, kapania2019hybrid}, \textit{gap acceptance} is a major factor that determines whether the pedestrian decides to cross or yield to the vehicle. It is defined as a time:
\begin{equation}
    t_{gap}=\frac{d_{front}}{v_{veh}}
\end{equation}
where $v_{veh}$ is the current vehicle velocity, and $d_{front}$ is the the distance to the interaction shown in Fig.~\ref{fig:scenario_crossing}. $t_{gap}$ is updated as time evolves and is compared with the threshold of the gap acceptance $\tau_{gap}$, which is drawn from a normal distribution $N(\mu_{gap}, \sigma_{gap})$. The statistics follows the results in~\cite{feliciani2017simulation}.

\subsection{Pedestrian Dynamics}
Instead of assuming a straight motion with constant crossing speed~\cite{chen2017evaluation, kapania2019hybrid}, the social force model applies a 2D point-mass Newtonian dynamics~\cite{yang2020social}: 
\begin{equation}
    \Ddot{x}_p=\frac{1}{m_p}f_{total},
\end{equation}
where $x_p\in\mathbb{R}^4$ is the pedestrian state vector that represents positions and velocities in x, y axes, respectively, $m_p$ is the mass of the pedestrian, and $f_{total}\in\mathbb{R}^2$ is the total applied force, which is detailed in the following subsection. In the simulation, the point-mass dynamics was discretized with the time step $\Delta t$. The dynamics also imposes constraints on the velocity $v_{p,max}$ and the acceleration $a_{p,max}$.

\subsection{Multi-State Social Force Interaction Model}
\subsubsection{Social Force}
Social force model describes each type of the interaction as a virtual force that applies on the pedestrian dynamics. The cumulative interaction effect is the summation of all individual virtual forces. In the proposed model, the total force was designed as:
\begin{equation}
    f_{total}=f_{des}+f_{veh},
    \label{eq:f_total}
\end{equation}
where $f_{des}\in\mathbb{R}^2$ is the destination force and $f_{veh}\in\mathbb{R}^2$ is the vehicle effect force. 

The destination force concretizes the pedestrian's desire to reach a particular location. Given a destination position, this force adjusts the pedestrian's velocity to walk toward the given destination with the desired speed. The destination force $f_{des}$ is defined as:
\begin{equation}
    f_{des}=k_{des}(v_p-v_d),
\end{equation}
where $v_p\in\mathbb{R}^2$ is the current pedestrian velocity vector, $v_d\in\mathbb{R}^2$ is the desired velocity vector that points to the destination, and $k_{des}$ is a scalar parameter that magnifies the difference between $v_p$ and $v_d$. The desired velocity vector is defined as $v_d:=v_0\cdot \frac{s_{des}-s_{p}}{\sqrt{|s_{des}-s_{p}|^2+(\sigma_{des})^2}}$. In the definition, $s_{des}\in\mathbb{R}^2$ is the destination, $s_p\in\mathbb{R}^2$ is the current pedestrian position, and $\sigma_{des}$ is a scalar parameter that decreases the desired speed when the pedestrian is getting close to the destination. $v_0$ is the desired speed magnitude that represents the most comfortable walking speed. $v_0$ is drawn from a normal distribution $N(\mu_{v_0}, \sigma_{v_0})$. The distribution follows the statistic results in~\cite{chandra2013speed}.

The vehicle effect is defined as a repulsive force:
\begin{equation}
    f_{veh}=A_{veh}\cdot \exp(-b\cdot d_{v2p})\cdot \vec{n}_{v2p}.
\end{equation}
$d_{v2p}$ is the distance from the influential point of the vehicle $s_{inf}$ to the current pedestrian position $s_p$. The influential point $s_{inf}$ is selected as the point on the vehicle contour that is closest to $s_p$. A pedestrian radius $R_p$ and an extension length $l_e$ for the contour are considered as the buffer. Therefore, $d_{v2p}:=|s_p-s_{inf}|-R_p-l_e$. $\vec{n}_{v2p}$ is a unit direction vector, pointing from $s_{inf}$ to $s_p$. The force magnitude applies an exponential relationship, with parameters $A_{veh}$ and $b_{veh}$. More details about the $f_{veh}$ can be found in~\cite{yang2020social}.

\subsubsection{State Transition}

\begin{figure}
    \centering
    \includegraphics[width=0.9\linewidth]{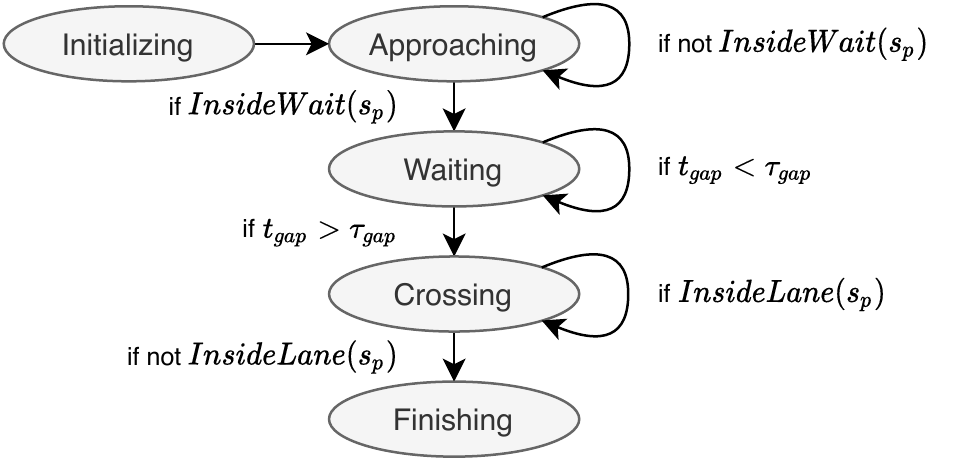}
    \caption{State transition of the social force model}
    \label{fig:state_transition}
\end{figure}

A crossing behavior is decomposed into several phases, as shown in Fig.~\ref{fig:state_transition}, during which the status of the social force model is slightly different:
\begin{itemize}
    \item \textit{Initializing}: The desired speed $v_0\sim N(\mu_{v_0}, \sigma_{v_0})$ and the gap threshold $\tau_{gap}\sim N(\mu_{gap}, \sigma_{gap})$ are obtained. The destination $s_{des}$ is set to be the waiting area in Fig.~\ref{fig:scenario_crossing}.
    
    \item \textit{Approaching}: Only the destination force $f_{des}$ is effective in equation~(\ref{eq:f_total}). If the waiting area, noted as \textit{InsideWait()} in Fig.~\ref{fig:state_transition}, is reached, switch to the next state.
    % The model is constantly checking if the waiting area is reached, as noted as \textit{InsideWait()} in Fig.~\ref{fig:state_transition}. If the waiting area is reached, the model switches to the next state.
    
    \item \textit{Waiting}: At this state, the pedestrian judges the gap $t_{gap}$. If $t_{gap}>\tau_{gap}$, the pedestrian starts to cross (by switching the destination $s_{des}$ to be the destination area in Fig.~\ref{fig:scenario_crossing}) and switches to the next state. Otherwise, the pedestrian just waits in the waiting area until $t_{gap}>\tau_{gap}$, which could be either the vehicle has passed the crossing position or the vehicle slows down and yields to the pedestrian. Still, only $f_{des}$ is effective.
    
    \item \textit{Crossing}: Pedestrian is within the lane where the vehicle is driving. Both the destination force $f_{des}$ and the vehicle effect force $f_{veh}$ are effective. 
    % In the middle of the crossing, if it is necessary for the pedestrian to accelerate (for example, when the vehicle executes aggressive maneuvers as if the vehicle doesn't see the pedestrian), 
    The desired speed $v_0$ will be temporarily changed if the pedestrian needs to avoid aggressive vehicle maneuvers (e.g., not yielding but accelerating). This is achieved by comparing the time to collision (TTC) for the vehicle $t_{TTC}:=\frac{d_{front}}{v_{veh}}$ with the time to finishing the crossing (TTF) for the pedestrian $t_{TTF}:=\frac{d_{rem}}{v_0}$, where $d_{rem}$ is the remaining distance from current pedestrian position to the other edge of vehicle driving lane. If $t_{TTC}<t_{TTF}$, then the updated desired speed $v_0'=d_{rem}/t_{TTC}$. Once the pedestrian leaves the vehicle driving lane, noted as $InsideLane()$ in Fig.~\ref{fig:state_transition}, switch to the next state.
    
    \item \textit{Finishing}: The pedestrian continues to the destination area. The vehicle effect force $f_{veh}$ is no longer effective.
\end{itemize}

% --------------------------------------------------------------------------------
\section{VEHICLE MOTION}
% The vehicle motion is governed by a longitudinal controller that considers the predicted pedestrian motion in a certain time horizon. 

\subsection{Pedestrian Motion Prediction}

Pedestrian motion prediction usually applies a system $T_{pred}=f_{pred}(T_{obs})$ that inputs an observed trajectory $T_{obs}=\{x_p(k-M+1),x_p(k-M+2),\cdots,x_p(k)\}$ of length $M$ and outputs a predicted trajectory $T_{pred}=\{x_p(k+1),x_p(k+2),\cdots,x_p(k+N)\}$ of length $N$. This work applies a linear pedestrian motion predictor, in which only the last observed pedestrian state $x_p(k)$ is used as input, and the output states are propagated by assuming the same velocity last observed.

% This work assumes the complete knowledge of the pedestrian state $x_p(k)$ at the current time step $k$ as well as the history trajectory $x_p(j), \forall j<k$. The fundamental problem of pedestrian motion prediction is formulated as a system that inputs an observed history trajectory and outputs a predicted future trajectory: $T_{pred}=f_{pred}(T_{obs})$, where $T_{obs}=\{x_p(k-M+1),x_p(k-M+2),\cdots,x_p(k)\}$ is the observed pedestrian states of past $M$ steps and $T_{pred}=\{x_p(k+1),x_p(k+2),\cdots,x_p(k+N)\}$ is the predicted pedestrian states of future $N$ steps. In this work, a simple linear pedestrian motion predictor was applied, in which only the last observed pedestrian state $x_p(k)$ is used as input, and the output states are propagated by assuming a constant velocity that uses the last observed one. 

% consider adding an equation here: is it necessary?
% More advanced pedestrian motion predictors can be applied as long as they satisfy the proposed structure. Other information such as the scene context and the states of the surrounding road users can also be considered to improve the accuracy of the pedestrian motion predictor. % citation required.

\subsection{Vehicle Dynamics}

A longitudinal point-mass model with drag effect is applied as the vehicle dynamics: $M\Ddot{s}(t)+\alpha\Dot{s}(t)=u(t)$, where $s$ is the longitudinal position, $M$ is the vehicle mass, $\alpha$ is a drag coefficient, and $u$ is the control action (throttle/brake). Rewriting the dynamics using a state vector $x=[s,\dot{s}]^T$ (position and velocity) into discretized time of $\Delta t$, we have:
\begin{equation}
    \label{eq:vehicle_dynamics}
    x(k+1) = Ax(k)+Bu(k)
\end{equation}
where $ A = 
    \begin{bmatrix}
        1 & \Delta t \\
        0 & 1-\frac{\alpha\Delta t}{M}
    \end{bmatrix},
    B = 
    \begin{bmatrix}
        0 \\
        \Delta t
    \end{bmatrix} $, 
and $u(k)$ is the discretized control action. A constraint $[v_{min}, v_{max}]$ on speed is imposed. Also, the control action is bounded by $[u_{min},u_{max}]$ and the control action rate is bounded by $[\Delta u_{min},\Delta u_{max}]$.

\subsection{Control Strategies}

The speed controller has two objectives: (a) keeping a safe distance to the pedestrian; (b) maintaining the desired speed as much as possible. This work analyzed 3 different control strategies. A vanilla pure velocity keeping control (VKC) was implemented as a baseline and to test the pedestrian behavior under extreme conditions as well. It is compared with an obstacle avoidance control (OAC) and a model predictive control (MPC) that use the predicted pedestrian motion.

\subsubsection{Velocity Keeping Control (VKC)}
It simply applies a PI controller to keep the vehicle's desired velocity $v_{0,veh}$:
\begin{equation}
    u=K_P\cdot(v_{0,veh}-v_{veh})+K_I\cdot I,
\end{equation}
where $I(k)=\sum_{i=0}^k\big(v_{0,veh}(i)-v_{veh}(i)\big)$ is the cumulative error of the speed deviation until current time step $k$ and $K_P$, $K_I$ are the proportional gain and the integral gain, respectively.
\subsubsection{Obstacle Avoidance Control (OAC)}
It extends the VKC by adding a strategy to decelerate if the predicted pedestrian trajectory at any time obstructs the vehicle driving lane. In that case, the control action is obtained by:
% solving $\frac{v_{veh}+0}{2}\cdot \frac{v_{veh}}{u}=d_{front}-d_{safe}$:
\begin{equation}
    u=-\frac{v_{veh}^2}{2\cdot(d_{front}-d_{safe})},
\end{equation}
where $d_{safe}$ is the safe distance, which equals to the $d_{front}$ when the vehicle decelerates and stops right in front of the pedestrian with the minimum deceleration.

\subsubsection{Model Predictive Control (MPC)}
MPC predicts $N_p$ steps of the vehicle motion. 
Two safety criteria were designed and imposed on the constraints of the MPC problem.
% To avoid the collision with the pedestrian, a safety criterion was designed and imposed into the constraints of the MPC problem. The safety criteria are modeled in two aspects. 
First, for any predicted pedestrian state $x_p(k+m),m\in{1,2,\cdots,M}$ that lies in the vehicle driving lane, its longitudinal position along the road $s_{obs}(k+n),n\in\mathbb{N}$ is used as a longitudinal displacement constraint for the MPC. $\mathbb{N}\subseteq\{1,2,\cdots,N_p\}$ is the index set of the future time steps when the predicted $x_p(k+m)$ lies in the vehicle driving lane. It must satisfy
\begin{equation}
    |s_{obj}(k+n)-s(k+n)|>d_{safe},\forall n \in\mathbb{N},
    \label{eq:con_dis}
\end{equation}
where $d_{safe}$ is the safe distance that should be always kept between the vehicle and the pedestrian and $s(k+n)$ is the predicted vehicle longitudinal position at time step $k+n$. These constraints guarantee that the vehicle never collides with the pedestrian within the prediction horizon. 
Second, if the predicted pedestrian position at final step $N_p$ also lies in the vehicle driving lane, a constraint on the vehicle's last predicted speed should be added such that the vehicle is expected to stop in front of the pedestrian at least a safe distance of $d_{safe}$.
Therefore, a deceleration distance $d_{dec}:=\frac{v(k+N_p)}{2}\cdot\frac{v(k+N_p)}{|u_{min}|}$ that allows the vehicle to decelerate from the terminal speed to zero speed within maximum deceleration (i.e., minimum control action $u_{min}$, which is negative) is added to $d_{safe}$. However, the square of $v(k+N_p)$ is not supported as constraints in most available MPC solvers. We relax the constraint in a way such that $d_{dec}=\frac{v(k+N_p)}{2}\cdot\frac{v(k+N_p)}{|u_{min}|}\le \frac{v(k+N_p)}{2}\cdot \frac{v_{max}}{|u_{min}|}$ hence the constraint becoming linear, where $v_{max}$ is the maximum allowed speed. So, the terminal constraint is:
\begin{equation}
    |s_{obj}(k+N_p)-s(k+N_p)|>\frac{v_{max}}{2|u_{min}|}\cdot v(k+N_p) + d_{safe}.
    \label{eq:con_terminal}
\end{equation}
Naturally, constraints on the velocity, control action, and control action rate should be added, they are:
\begin{align}
    v_{min} & <v(k+n)<v_{max}, \forall n\in\{1,2,\cdots, N_p\} \label{eq:con_vel}\\ 
    u_{min} & <u(k+n)<u_{max}, \forall n\in\{0,1,\cdots, N_p-1\} \label{eq:con_acc} \\ 
    \Delta u_{min} & <\Delta u(k+n)<\Delta u_{max}, \forall n\in\{0,1,\cdots, N_p-1\} \label{eq:con_dacc}
\end{align}
Finally, the MPC problem is formulated as:
\begin{align}
    \boldsymbol{U}^*&=\underset{\boldsymbol{U}}{\arg\min}\Big(
        \sum_{n=1}^{N_p}w_v\big(v(k+n)-v_{0,veh}\big)^2 + 
        \sum_{n=0}^{N_p-1}w_u\big(u(k+n)\big)^2 
    \Big)
    \label{eq:QP_problem} \\
    &\text{s.t.}  \ \ 
    [s(k), v(k)]^T=x(k), \text{and}\ 
    (\ref{eq:vehicle_dynamics}) 
    (\ref{eq:con_dis})
    (\ref{eq:con_terminal})
    (\ref{eq:con_vel})
    (\ref{eq:con_acc})
    (\ref{eq:con_dacc})
    \notag,
\end{align}
where $x(k)$ is the current vehicle state, and $w_v$, $w_u$ are the weights for the cost of velocity and control, respectively. In extreme cases if solving the MPC fails, maximum deceleration $u_{min}$ is applied, with the constraint (\ref{eq:con_dacc}) still valid.

% --------------------------------------------------------------------------------

\begin{table}[h]
    \centering
    \caption{Parameters in the Simulation}
    \begin{tabular}{ccc|ccc}
        Symbol & Value & Units & Symbol & Value & Units\\
        \hline
        \hline
        $v_{p,max}$ & 2.5 & $m/s$   & $m_p$ & 80.0 & $kg$ \\ 
        $a_{p,max}$ & 5.0 & $m/s^2$ & $R_p$ & 0.27 & $m$ \\
        $\mu_{v_0}$    & 1.4 & $m/s$ & $\mu_{gap}$    & 2.5 & sec\\
        $\sigma_{v_0}$ & 0.2 & $m/s$ & $\sigma_{gap}$ & 4.0 & sec\\
        $A_{veh}$ & 200.0 & - & $\sigma_{des}$ & 1.0   & - \\
        $b_{veh}$ & 2.6   & - & $k_{des}$      & 300.0 & - \\
        \hline
        $M$ & 2000.0 & $kg$ & $\alpha$ & 100 & - \\
        $u_{min}$ & -7.0 & $m/s^2$ & $v_{min}$ & 0.0 & $m/s$\\
        $u_{max}$ & 7.0  & $m/s^2$ & $v_{max}$ & 22.5 & $m/s$\\
        $\Delta u_{min}$ & -5.0 & $m/s^3$ & $d_{safe}$ & 3.0 & $m$ \\
        $\Delta u_{max}$ & 5.0  & $m/s^3$ & $N_{pred}$ & 15 & - \\
        $K_P$ & 1.0 & - & $w_v$ & 1.0 & - \\
        $K_I$ & 0.1 & - & $w_u$ & 1.0 & - \\
    \end{tabular}
    \label{tab:parameters}
\end{table}

\section{EXPERIMENTS}

% A simulation program was created from scratch in Python to represent the scenario in Fig.~\ref{fig:scenario_crossing}.
A two-lane road was created for simulation experiments, with lane width $W_{lane}=3.2m$. 
% The vehicle is driving on the right lane. 
For each episode, the pedestrian was initialized at 2 meters from the right to the right edge of the road. The destination was set at 3.6 meters from the left to the other edge of the road. The waiting area was centered at 0.5 meters to the right road edge. 
These 3 points have the same longitudinal position. 
The vehicle was initialized with 30 different combinations of initial longitudinal position $d_{front,0}\in\{11.5,16.5,21.5,26.5,31.5,36.5\}$ and speed $\dot{s}_0\in\{2,4,6,8,10\}$ so that the majority of scenarios of different time gaps were covered. The desired speed was set as $v_{0,veh}=\dot{s}_0$. For each combination, 3 different control strategies were simulated for 200 times, respectively. CVXPY~\cite{diamond2016cvxpy} was applied as the MPC solver. Table~\ref{tab:parameters} shows the parameter values, which were manually tuned according to the parameter values in previous works~\cite{yang2020social, liu2015predictive, kapania2019hybrid}.
% The parameters were tuned by trial and error so that the performance of the overall system was verified by the human experience. Some parameters that are related to the social force model or the model predictive control also came from the previous practice in tuning the parameters in individual models.

% --------------------------------------------------------------------------------

\section{RESULTS}

In general, there are 4 hyper-parameters in the simulation configuration. They are the threshold of the accepted gap $\tau_{gap}$, the desired pedestrian walking speed $v_0$, the initial vehicle longitudinal position $d_{front}$, and the initial/desired vehicle speed $\dot{s}=v_{0,veh}$. The results were evaluated based on selected combinations of the above hyper-parameters.

\begin{figure}
    \centering
    \includegraphics[width=\linewidth]{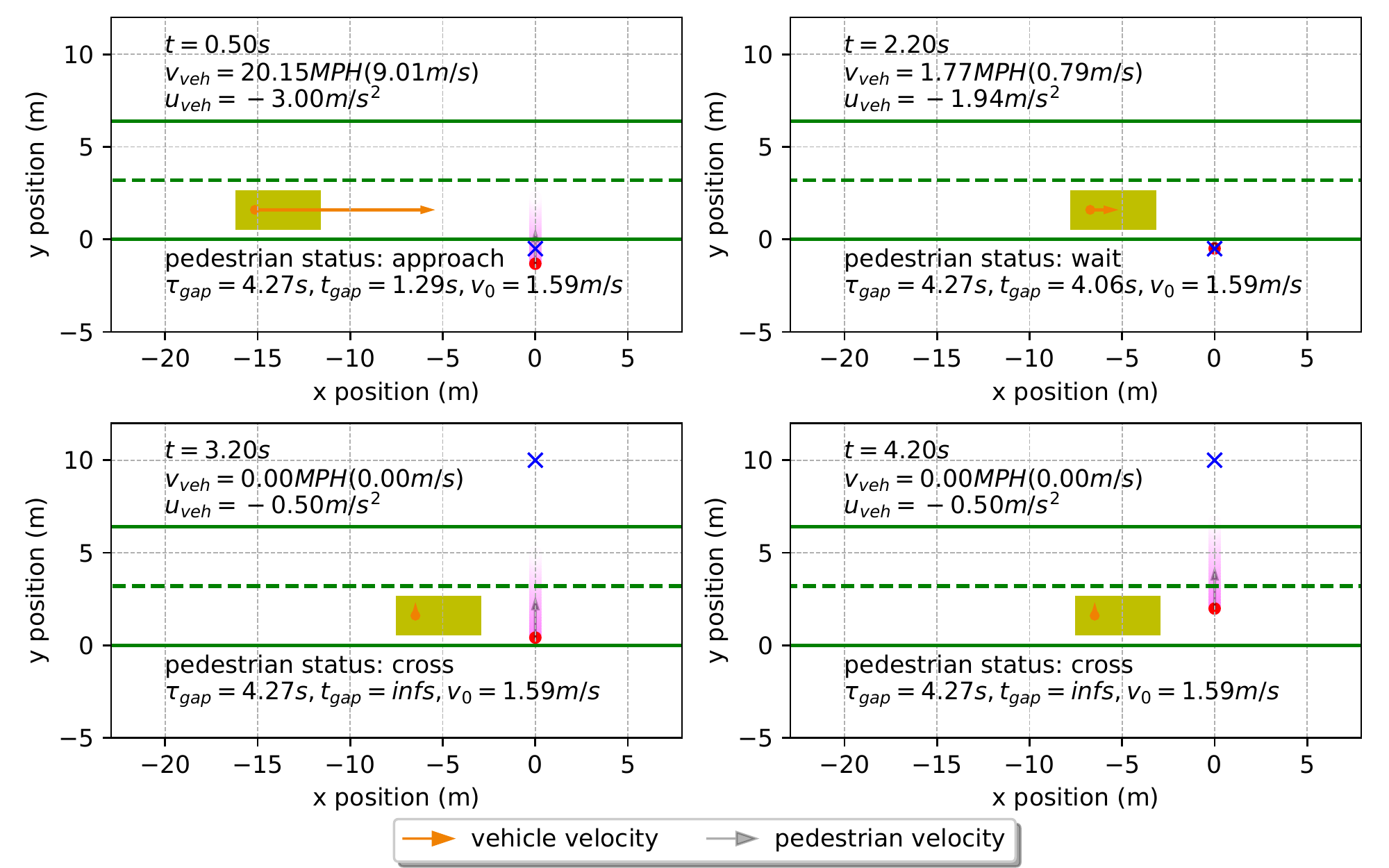}
    \caption{Screenshots of a simulation that applies MPC. The initial state of the vehicle (yellow box) is $d_{front,0}=16.5m$ and initial/desired speed $\dot{s}_0=10m/s$. The pedestrian (red circle) was initialized with a gap acceptance threshold of $\tau_{gap}=4.27s$ and a desired walking velocity $v_0=1.59m/s$. The blue 'x' is the pedestrian's destination. Purple shadows indicate the predicted pedestrian positions (lighter color means longer predicted time).}
    \label{fig:qualitative_scene}
\end{figure}

\begin{figure}
    \centering
    \includegraphics[width=\linewidth]{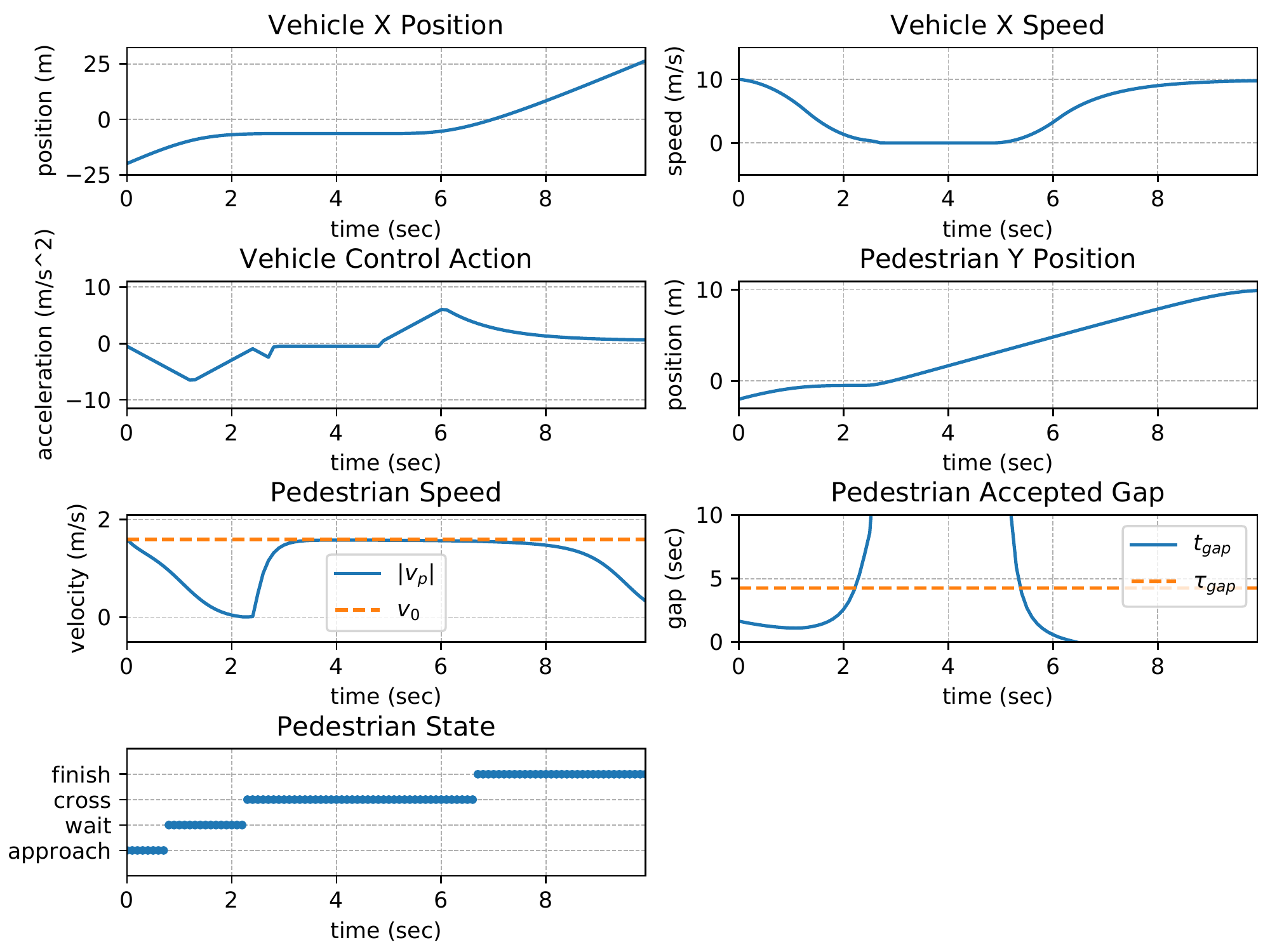}
    \caption{Evolution of some states that corresponds to the simulation in Fig.~\ref{fig:qualitative_scene}: vehicle's longitudinal (x) position, velocity, and control action; pedestrian's crossing distance (y position), speed, state, and the estimated gap.}
    \label{fig:qualitative_data}
\end{figure}

\subsection{Qualitative Evaluation}

Fig.~\ref{fig:qualitative_scene} shows the screenshots of a simulation using MPC. The evolution of the corresponding pedestrian state and vehicle state are plotted in Fig.~\ref{fig:qualitative_data}. At the beginning stage, the vehicle recognized the pedestrian who was approaching the edge of the road. The pedestrian motion predictor provided a sequence of predicted future positions that lie in the vehicle driving lane. Therefore, the MPC was generating the control action to slow down the vehicle. In the meantime, the pedestrian was also slowing down because $t_{gap}<\tau_{gap}$. As the vehicle was continuing to slow down, the estimated gap $t_{gap}$ was increasing (see the gap value in Fig.~\ref{fig:qualitative_data}), and around $t=2.2s$, $t_{gap}>\tau_{gap}$, which made the pedestrian switch from \textit{waiting} state to \textit{crossing} state. After that, the pedestrian was crossing the road while the vehicle was fully stopped and waiting for the pedestrian to complete the crossing. This example demonstrated a successful interactive VPI, which validated the effectiveness of the proposed framework.
% If $\tau_{gap}$ was initialized with a relatively large value, the result could be the pedestrian was yielding to the vehicle during the interaction.

\subsection{Quantitative Evaluation}

\begin{figure*}
    \centering
    \includegraphics[width=\textwidth]{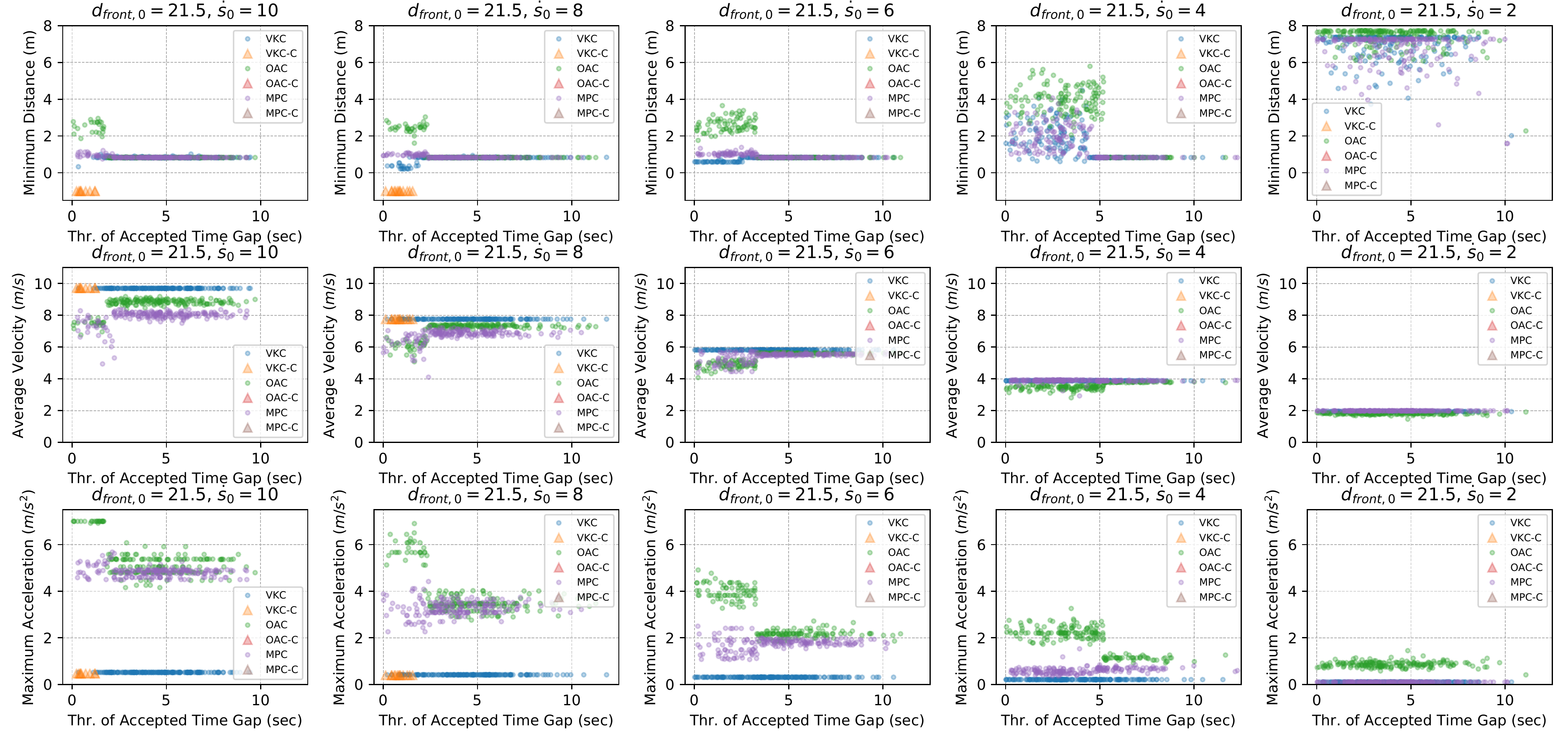}
    \caption{Quantitative analysis of the simulation results of $d_{front,0}=21.5$ and $\dot{s}_0\in\{2,4,6,8,10\}$. The 1st row shows the minimum distance from the pedestrian to any point of the vehicle contour during the entire simulation, the 2nd row shows the average velocity, and the 3rd row shows the maximum value of the absolute acceleration/deceleration. Each plot shows the results of running each control strategy for 200 times. In each plot, VKC indicates velocity keeping control, OAC indicates obstacle avoidance control, and MPC indicates model predictive control. The suffix -C indicates collision results.}
    \label{fig:quantitative_results}
\end{figure*}

The quantitative evaluation of the proposed framework was conducted to inspect the safety, efficiency, and smoothness of the vehicle, which follows the evaluation process in the work~\cite{kapania2019hybrid}. Fig.~\ref{fig:quantitative_results} shows the results of combinations of $d_{front,0}=21.5$ and all vehicle initial/desired speeds $\dot{s}_0$. The variance of the results corresponding to particular $\tau_{gap}$s was due to the variance of the pedestrian's desired speed $v_0$.

\subsubsection{Safety}
The minimum distance between the pedestrian and the vehicle during an entire episode is used to evaluate pedestrian safety. According to all the plots in the 1st row, if the threshold of $\tau_{gap}$ was larger than the gap estimated at the time when the vehicle was initialized, most likely the pedestrian decided to wait and yield to the vehicle. In these cases, the minimum distances were almost the same, which is approximately equal to the distance from the road edge to the vehicle's right side. For the cases of pedestrian not yielding, the OAC was comparatively safer than the other two (but at the expense of smoothness). Note that for the VKC, collision happened in some cases when the vehicle's initial/desired speed $v_{0,veh}$ is relatively high.

\subsubsection{Efficiency}
The average vehicle velocity is used to evaluate the driving efficiency. The VKC had the highest efficiency because it didn't react to the pedestrian at all, but the collision is unacceptable. Comparing the OAC with the MPC, the OAC outperformed the MPC at high speed $v_{0,veh}$, but was outperformed by the MPC as the $v_{0,veh}$ decreases.

\subsubsection{Smoothness}
The minimum value of the absolute acceleration/deceleration is used to evaluate the smoothness. Similarly to the efficiency, the VKC maintained the same minimum acceleration but collision happened. The MPC was always smoother than the OAC in general. 

In sum, considering safety, efficiency, and smoothness simultaneously, when the vehicle speed $v_{0,veh}$ is relatively low, the MPC outperformed the OAC and the VKC. This is because the MPC predicts the future vehicle motion so that the control action is optimized to consider both the safety, efficiency, and smoothness. And when $v_{0,veh}$ is low, the MPC has more allowed time to optimize its control action. The results associated with other $d_{front,0}$ also follow a similar pattern as illustrated in Fig.~\ref{fig:quantitative_results}.

% --------------------------------------------------------------------------------
\section{CONCLUSIONS}
This work proposed a framework to address the VPI in uncontrolled crossing scenarios. A novel multi-state social force pedestrian motion model was integrated into the framework. The behavior of both the vehicle and the pedestrian in the experiments of 3 different vehicle control strategies demonstrated the effectiveness of the proposed framework. 

Major further work would be improving the pedestrian model by considering demographics and by leveraging VPI datasets. Framework components such as pedestrian motion predictor could also be replaced with a more accurate one. In terms of vehicle control, the next step would be better integrating pedestrian uncertainty into the controllers. And lastly, of course, the framework should be extended to cover more VPI scenarios.


%%%%%%%%%%%%%%%%%%%%%%%%%%%%%%%%%%%%%%%%%%%%%%%%%%%%%%%%%%%%%%%%%%%%%%%%%%%%%%%%

%%%%%%%%%%%%%%%%%%%%%%%%%%%%%%%%%%%%%%%%%%%%%%%%%%%%%%%%%%%%%%%%%%%%%%%%%%%%%%%%



% \section*{ACKNOWLEDGMENT}

% Thanks to xxxxx

%%%%%%%%%%%%%%%%%%%%%%%%%%%%%%%%%%%%%%%%%%%%%%%%%%%%%%%%%%%%%%%%%%%%%%%%%%%%%%%%

\bibliographystyle{IEEEtran}
\bibliography{IEEEabrv, mybib}

% \begin{thebibliography}{99}
% \end{thebibliography}

\end{document}